\documentclass[runningheads,a4paper]{llncs}

\usepackage{amssymb}
\usepackage{graphicx}
\usepackage{caption, booktabs}
\usepackage{subcaption}
\usepackage{listings}
\usepackage{xcolor}
\usepackage{algorithm2e}
\usepackage{hyperref}
\usepackage[para]{footmisc}

\usepackage{url}
\urldef{\mailsa}\path|{alfred.hofmann, ursula.barth, ingrid.haas, frank.holzwarth,|
\urldef{\mailsb}\path|anna.kramer, leonie.kunz, christine.reiss, nicole.sator,|
\urldef{\mailsc}\path|erika.siebert-cole, peter.strasser, lncs}@springer.com|    
\newcommand{\keywords}[1]{\par\addvspace\baselineskip
\noindent\keywordname\enspace\ignorespaces#1}

\begin{document}


\definecolor{forestgreen}{RGB}{34,139,34}
\definecolor{orangered}{RGB}{239,134,64}
\definecolor{darkblue}{rgb}{0.0,0.0,0.6}
\definecolor{gray}{rgb}{0.4,0.4,0.4}
\definecolor{codegreen}{rgb}{0,0.6,0}
\definecolor{codegray}{rgb}{0.5,0.5,0.5}
\definecolor{codepurple}{rgb}{0.58,0,0.82}
\definecolor{backcolour}{rgb}{0.95,0.95,0.92}

\lstdefinestyle{XML} {
    language=XML,
    extendedchars=true, 
    breaklines=true,
    breakatwhitespace=true,
    emph={},
    emphstyle=\color{red},
    basicstyle=\ttfamily,
    columns=fullflexible,
    commentstyle=\color{gray}\upshape,
    morestring=[b]",
    morecomment=[s]{<?}{?>},
    morecomment=[s][\color{forestgreen}]{<!--}{-->},
    keywordstyle=\color{orangered},
    stringstyle=\ttfamily\color{black}\normalfont,
    tagstyle=\color{darkblue}\bf,
    morekeywords={attribute,xmlns,version,type,release},
    otherkeywords={attribute=, xmlns=},
}

\lstdefinelanguage{PDDL}
{
  sensitive=false,    
  morecomment=[l]{;}, 
  alsoletter={:,-},   
  morekeywords={
    define,domain,problem,not,and,or,when,forall,exists,either,
    :domain,:requirements,:types,:objects,:constants,
    :predicates,:action,:parameters,:precondition,:effect,
    :fluents,:primary-effect,:side-effect,:init,:goal,
    :strips,:adl,:equality,:typing,:conditional-effects,
    :negative-preconditions,:disjunctive-preconditions,
    :existential-preconditions,:universal-preconditions,:quantified-preconditions,
    :functions,assign,increase,decrease,scale-up,scale-down,
    :metric,minimize,maximize,
    :durative-actions,:duration-inequalities,:continuous-effects,
    :durative-action,:duration,:condition
  }
}

\lstdefinelanguage{Srv}{
keywords = [1]{bool, uint8, int32, uint64, float32, float64, string, Header, Point, Quaternion, time},
comment=[l]{\#}
}

\lstdefinestyle{mystyle}{
  backgroundcolor=\color{backcolour},   commentstyle=\color{codegray},
  keywordstyle=\color{codegreen},
  numberstyle=\tiny\color{codegray},
  stringstyle=\color{codepurple},
  basicstyle=\ttfamily\footnotesize,
  breakatwhitespace=false,         
  breaklines=true,                 
  captionpos=b,                    
  keepspaces=true,                 
  numbers=left,                    
  numbersep=5pt,                  
  showspaces=false,                
  showstringspaces=false,
  showtabs=false,                  
  tabsize=2
}
\lstset{style=mystyle}

\mainmatter  

\title{Simplifying Robot Programming using Augmented Reality and End-User Development}

\titlerunning{Augmented Reality-Assisted Robot Programming}

%
%
%

\author{Enes Yigitbas \and Ivan Jovanovikj \and Gregor Engels}

\authorrunning{E. Yigitbas et al.}

\institute{Paderborn University\\ Zukunftsmeile 2, 33102 Paderborn, Germany\\
\email{firstname.lastname@upb.de}, 
}

%
%

\toctitle{Lecture Notes in Computer Science}
\tocauthor{Authors' Instructions}
\maketitle

\begin{abstract}
Robots are widespread across diverse application contexts. Teaching robots to perform tasks, in their respective contexts, demands a high domain and programming expertise. However, robot programming faces high entry barriers due to the complexity of robot programming itself. Even for experts robot programming is a cumbersome and error-prone task where faulty robot programs can be created, causing damage when being executed on a real robot. To simplify the process of robot programming, we combine Augmented Reality (AR) with principles of end-user development. By combining them, the real environment is extended with useful virtual artifacts that can enable experts as well as non-professionals to perform complex robot programming tasks. Therefore, Simple Programming Environment in Augmented Reality with Enhanced Debugging (SPEARED) was developed as a prototype for an AR-assisted robot programming environment. SPEARED makes use of AR to project a robot as well as a programming environment onto the target working space. To evaluate our approach, expert interviews with domain experts from the area of industrial automation, robotics, and AR were performed. The experts agreed that SPEARED has the potential to enrich and ease current robot programming processes.
\end{abstract}

\section{Introduction}\label{sec:intro}
Robots are becoming ubiquitous and they are used nowadays in different settings with typical application domains like education, household, or industry. They come in diverse forms and shapes depending on the task they were designed for. In the manufacturing industry, an increasing number of robots are used for assembly tasks such as screwing, welding, painting or cutting~\cite{DBLP:journals/bise/LasiFKFH14}.

Tasks are domain-dependent as different contexts of use demand different precision/safety regards (i.e., educational robots in school vs. robots in the manufacturing industry). Describing a task requires process knowledge as well as domain knowledge. Transferring this as an executable movement to the robot demands (robot) programming proficiency. However, robot programming is a complex and time-consuming task, where programming errors may occur, e.g., minor offsets in coordinates or differences between the test environment and real environment~\cite{DBLP:conf/ismar/PettersenPSEL03}. On a real robot, these may cause hardware damage up to irrecoverable damage to a person. Thus, the reduction of errors and their prevention is important. Consequently, different challenges exist which cause the high complexity of robot programming. Based on an extensive literature research (see Section 2) related to the topics of AR-assisted robot programming and robot programming by demonstration, we have identified the following major challenges:

\textbf{C1 Program state visualization:}
In robot programming, the visualization of the robot and its parameters at a current state can be challenging.
For validation purposes and to simplify trajectory programming an intuitive visualization of context information concerning path trajectory, the goal of the next movement, and end-effector status is required (\cite{DBLP:conf/icse/ShepherdKF19,DBLP:conf/vw/FangON09,Fang2012}). Developers need to be able to validate whether the execution of the current program causes the expected effect. For supporting robot programmers in identifying the current state of the robot and possible configuration errors, feedback about current path trajectories or states of different robot parts is required.

\textbf{C2 Root cause analysis:}
Enabling a correct root cause analysis of programming errors is important as logic errors may occur when programming~\cite{DBLP:conf/icawst/YoshizawaW18}. Such errors can lead to unintended actions in the robot behavior. Hence, a common challenge robot programmers need help with, is finding root causes for these failures. Problems in source code can be solved if the robot programmer is able to identify these causes for a specific problem, e.g., imprecise coordinates or wrong end-effector state at a specific code point.

\textbf{C3 Vendor robot programming language proficiency:}
The reduction of needed vendor robot programming language proficiency should also be addressed.
Here, the focus lies on approaches using code or code-alike representations of robot programs. Usually, every vendor has its own programming language~\cite{DBLP:conf/isr/PanPLDN10}. Being manufacturer-independent reduces barriers to integration~\cite{Lambrecht2011}. Furthermore, it removes the need to learn a new programming language for every robot vendor. Thus, it probably reduces the mental-load when programming and also allows vendor-independent, uniform programming.

\textbf{C4 3-Dimensional Thinking:}
The correct determination of 3-dimensional data is important as robots operate in their own 3-dimensional space. A correct interpretation of coordinates and their translation from/into source code is necessary to create working programs. The interpretation can be hard because parts of the robot may have different coordinate origins than the real world. Translating and mapping them to each other is not trivial. Especially 3D models represented on 2D Screens cause issues in positioning because of missing depth~\cite{DBLP:conf/icse/ShepherdKF19}. Humans intuitively locate things in 3D-space and dynamically adapt their estimations as they get closer. Hence, an intuitive definition of targets or coordinates, as well as an understandable representation of (arbitrary) targets, are required for simplifying robot programming.

\textbf{C5 Environment-specific information:}
The correct representation of environmental constraints is also very important. For example, when an object the robot might want to grab could be a couple of centimeters off since the last development state. Furthermore, there could be physical obstacles in the working environment of the robot. Therefore, representing this kind of information, as fixtures~\cite{DBLP:journals/ras/AleottiCR04} or cues~\cite{DBLP:journals/tsmc/AleottiCR05}, as well as integrating them at development time, e.g., interacting with a virtual clone of the real object, should be addressed. This is especially important when the recreation of a full simulation for every difference ~\cite{DBLP:conf/taros/GianniFP13} is not possible or a considerable option. 

\textbf{C6 Modification of existing code:}
When working with existing source code bases of robot programs, it has to be ensured that functional modifications are causing the expected change in behavior. For supporting robot programmers in adjusting and adaptation of existing robot code it is important to import existing parts of the code and enable a modification in the same manner as the creation of new programs. This of course means to consider above-mentioned challenges and to reload the modified code to the robot so that the changes in the behavior can be reflected to the robot system.

\begin{figure}[hbt!]
\centering
\includegraphics[width=0.9\textwidth]{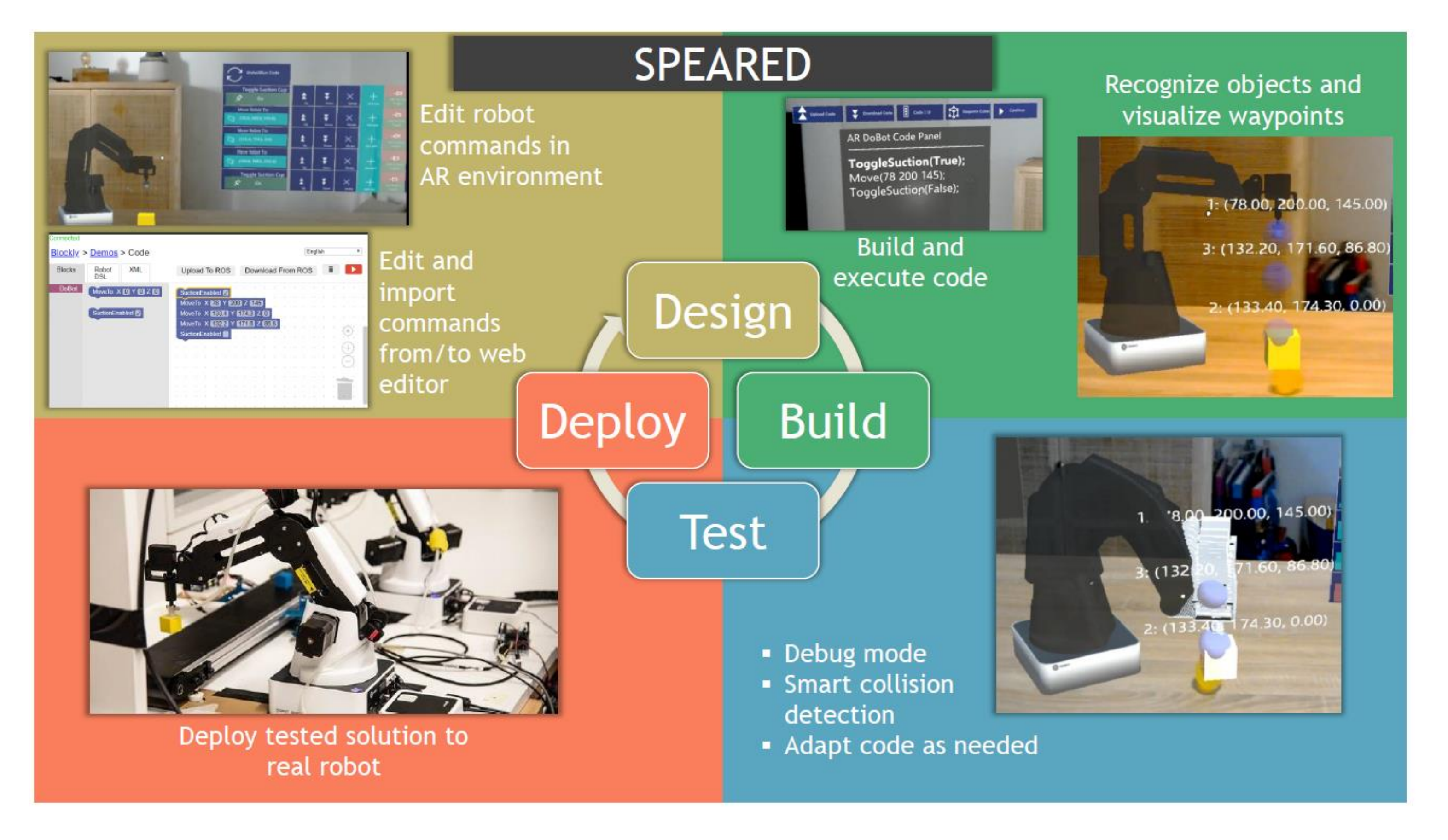}
\caption{Overview of the AR-assisted robot programming environment}
\label{fig:overview}
\end{figure}

To tackle the above-mentioned challenges, we have developed a novel AR-assisted robot programming environment called SPEARED. A general overview of the solution idea of SPEARED is depicted in Fig.\ref{fig:overview}. 

SPEARED is an AR-assisted robot programming environment that supports the different phases of the robot programming process, namely \textit{design}, \textit{build}, \textit{test}, and \textit{deploy} while addressing the aforementioned challenges. To address \textit{Challenge C1}, a head-mounted AR device is used which shows the robot model (current working state), detected objects, and goals of the next movement (refer to top-left and top-right screenshots of Fig.~\ref{fig:overview}).
Furthermore, the current robot code is shown (refer to the "AR DoBot Code Panel" in the top middle of Fig.~\ref{fig:overview} or refer to Fig.~\ref{fig:arDobotCodePanel}), so the user knows which code is executed and is able to find errors. Thus, tackling \textit{Challenge C2}.
To handle the \textit{Challenge C3}, the code in the "AR DoBot Code Panel" is written in a domain-specific language (DSL) that is vendor-independent. New code can be created, existing code adapted and both of them executed. This is done by using building blocks (e.g. move the robot to coordinate A, enable the end-effector of the robot). These building blocks have placeholders (i.e. the point A or whether the end-effector should be enabled or disabled).
To tackle \textit{Challenge C4}, these programming building blocks can also be set in the AR environment. The code component is visualized in the top-left image of Fig.~\ref{fig:overview}. The AR environment enables the previously mentioned code interactions as well as some quality of life features e.g. voice commands, transforming the User Interface (UI) (resize, rotate, move) based on gestures, etc.
To simplify programming inside the AR environment, the above mentioned simple and lightweight DSL allows the users to abstract from scenario-dependent settings (e.g. speed of movement, movement type). This block-based DSL enables the user to perform movement tasks as well as changing the state of the end-effector. Furthermore, specific interaction objects are emphasized (refer to the detected object/yellow cube in Fig.~\ref{fig:overview}). This enables the handling of \textit{Challenge C5}.
The aforementioned code creation and modification can be done in the AR application as well as in a non AR setting (refer to the top-left of Fig.~\ref{fig:overview}). The created artifacts can be synchronized in both directions. Here the translation to the aforementioned block-language enables a consistent programming experience for old and new code. Thus, it tackles \textit{Challenge C6}. The code being executed is sent to a simulator that moves the robot. These movements are then shown and subscribed to the model of the robot in the AR application. This also allows the user to visualize the current working state. The final code can be deployed to the real robot (see bottom left of Fig.~\ref{fig:overview}).
To evaluate the fulfillment of the challenges, the AR solution has been evaluated based on expert interviews from the domain of industrial automation, robotics, and AR.

The rest of the paper is structured as follows. Sec.~\ref{sec:related_work} introduces and discusses related work in the areas of robot programming and augmented reality. The conceptual solution and the implementation of the SPEARED framework are presented in Sec.~\ref{sec:conceptual_solution} and Sec.~\ref{sec:implementation}, respectively. In Sec.~\ref{sec:usability_evaluation}, we present and discuss the main results of the expert interviews. Finally, Sec.~\ref{sec:conclusion} presents conclusions and directions for future work.

\section{Related Work}\label{sec:related_work}

Augmented Reality (AR) and Virtual Reality (VR) have been a topic of intense research in the last decades. In the past few years, massive advances in affordable consumer hardware and accessible software frameworks are now bringing AR and VR to the masses. While VR interfaces support the interaction in an immersive computer generated 3D world and have been used in different application domains such as training~\cite{DBLP:conf/vrst/YigitbasJSE20}, robotics \cite{DBLP:journals/corr/abs-2103-10804}, education~\cite{DBLP:conf/mc/YigitbasTE20}, or healthcare~\cite{DBLP:conf/mc/YigitbasHE19}, AR enables the augmentation of real-world physical objects with virtual elements. In previous works, AR has been already applied for different aspects such as product configuration (e.g., \cite{DBLP:conf/hcse/GottschalkYSE20}, \cite{DBLP:conf/hcse/GottschalkYSE20a}), prototyping \cite{DBLP:conf/hcse/JovanovikjY0E20}, planning and measurements \cite{DBLP:conf/eics/EnesScaffolding} or for realizing smart interfaces (e.g., \cite{DBLP:conf/eics/KringsYJ0E20}, \cite{DBLP:conf/interact/YigitbasJ0E19}). Besides this broad view of application domains, in recent years, several approaches have addressed the problem of enabling non-programmers to program robots. Apart from classical programming approaches based on end-user development~\cite{DBLP:journals/vlc/CoronadoMIV20}, two major fields of related work in this direction are robot programming by demonstration and augmented reality-assisted robot programming approaches. 

\subsection{Robot programming by demonstration}\label{subsec:PbD}

\begin{figure}[hbt!]
\centering
\includegraphics[width=0.8\textwidth]{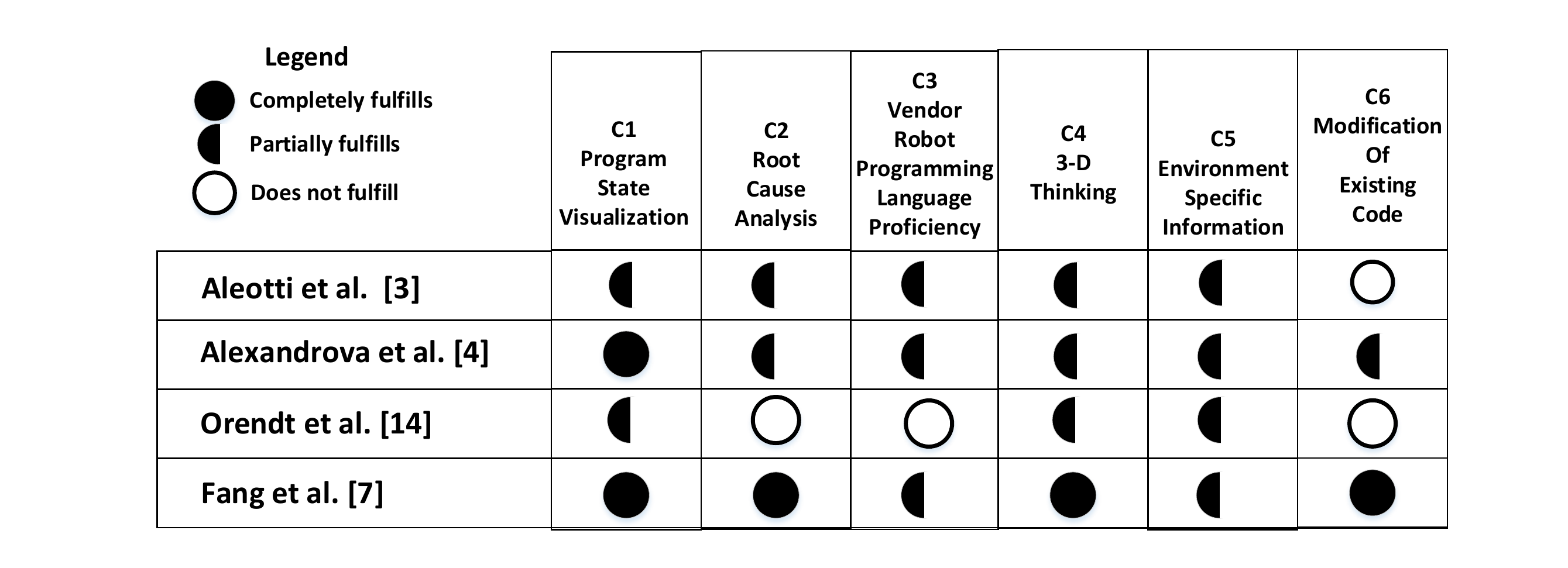}
\caption{Evaluation of PbD approaches}
\label{fig:relWorkPbD}
\end{figure}

Robot Programming by Demonstration (PbD), also known as imitation learning, is a widespread technique to teach robots how to perform tasks. Hence, PbD is used as an alternative or addition to traditional programming. In PbD, the resulting artifact is not necessarily a program that can be executed. It can be a more generic representation of the task presented, as in Aleotti et al.\cite{DBLP:conf/have/AleottiMC14}. Here, a visuo-haptic AR interface was used for robot programming by demonstration. The task \textit{lay the table} was taught to a robot. To be specific, the precedence relations that hold were learned e.g. the dinner plate has to be put on the table first, then the soup plate has to be put on top of it. These relations were learned via multiple demonstrations of the task. The demonstration was done with a haptic input device for performing the interaction and a camera-based AR system for visual feedback. 
A further PbD based approach is presented by Orendt et al. \cite{7745110}. In contrast to~\cite{DBLP:conf/have/AleottiMC14}, it enables the execution of the created artifact on the real robot. To be precise, they focused on one-shot learning by kinesthetic teaching. In other words, the task was demonstrated one time (one-shot) by moving the robot arm. 
Instead of using a real robot, a visual AR environment for PbD was used in Fang et al.\cite{Fang2014}. Here, a pick and place task and path following operations were taught by kinesthetic teaching. Therefore, a tracker marked cube was used as a demonstration instrument. The results were saved as sets of points with end-effector orientation. These were visualized after recording the task. Another PbD-based approach is presented in \cite{DBLP:conf/rss/AlexandrovaCHT14}, where Alexandrova et al. present a method to generalize from programming by demonstrations. They have developed a Graphical User Interface (GUI) which was able to edit actions after they were demonstrated. In total, as depicted in Fig.\ref{fig:relWorkPbD}, we can see that existing PbD approaches do not fully cover the mentioned challenges \textit{C1} - \textit{C6}. Still, the presented works offer insight about relevant approaches and their upsides and downsides: One-shot learning offers high intuitiveness, but does often lack the integration of adaptation capabilities. Having these integrated, one-shot learning as well as using multiple demonstrations allows tackling the challenges \textit{C1} and \textit{C2}. Using visual depth information, or force-feedback as in~\cite{DBLP:conf/have/AleottiMC14}, by utilizing AR tackles \textit{Challenge C4} and enables a possible integration of environment-specific information (\textit{Challenge C5}). PbD often relies on environments that defer from using a programming language. Thus, \textit{Challenge C3} is often missing by design because a code-alike representation itself is missing. In conclusion, existing PbD approaches do not tackle all challenges to serve as a solution for this work. 

\subsection{Augmented Reality-assisted robot programming}\label{subsec:arbot}

In the following, we briefly present and evaluate related approaches that follow the idea of augmented reality-assisted robot programming. 

Shepherd et al. \cite{DBLP:conf/icse/ShepherdKF19} used a video-based AR approach for robot programming. A block-based Integrated Development Environment (IDE) was embedded onto the screen. The block-based IDE is a CoBlox derivate. CoBlox is designed for offline robot programming and normally features a robot and environment simulation (\cite{8120406}, \cite{DBLP:conf/scam/ShepherdFWFLA18}, \cite{DBLP:conf/chi/WeintropASFLSF18}). However, in contrast to the simulation, in this paper, a real robot was used. Additionally, waypoints, showing the current planned path, were projected in the AR environment. In \cite{DBLP:conf/icse/ShepherdKF19} it was also noted, that programming with controllers as well as having to move the real robot with hands is cumbersome. Another approach is to use hands for both interactions. This approach was used in Gadre et al. \cite{DBLP:conf/icra/GadreRCPTK19} where Mixed Reality (MR) interface was proposed for creating and editing waypoints. The created waypoints could be grouped, the resulting action previewed, and the resulting program executed on the real robot. The solution was tested against a monitor interface in a usability study using Microsoft HoloLens. Ong et al. \cite{ong} proposed an AR-assisted robot programming system that allows  users to program pick-and-place as well as welding tasks by demonstrating the task/path. These paths can be selected by showing the full path, showing the start and endpoints, or selecting features based on Computer-Aided Design (CAD), e.g., edges. The robot motion was simulated and augmented in AR. Rosen et al. \cite{DBLP:journals/ijrr/RosenWPCTKT19} proposed an MR interface for programming robots. Here a manipulation of the movement starting point and goal, as well as end-effector orientation, via hand interaction, is possible. The AR device used is a Microsoft HoloLens. MoveIt is the motion planning tool used to calculate a path between the two aforementioned points (start point and endpoint). 

\begin{figure}[hbt!]
\centering
\includegraphics[width=0.8\textwidth]{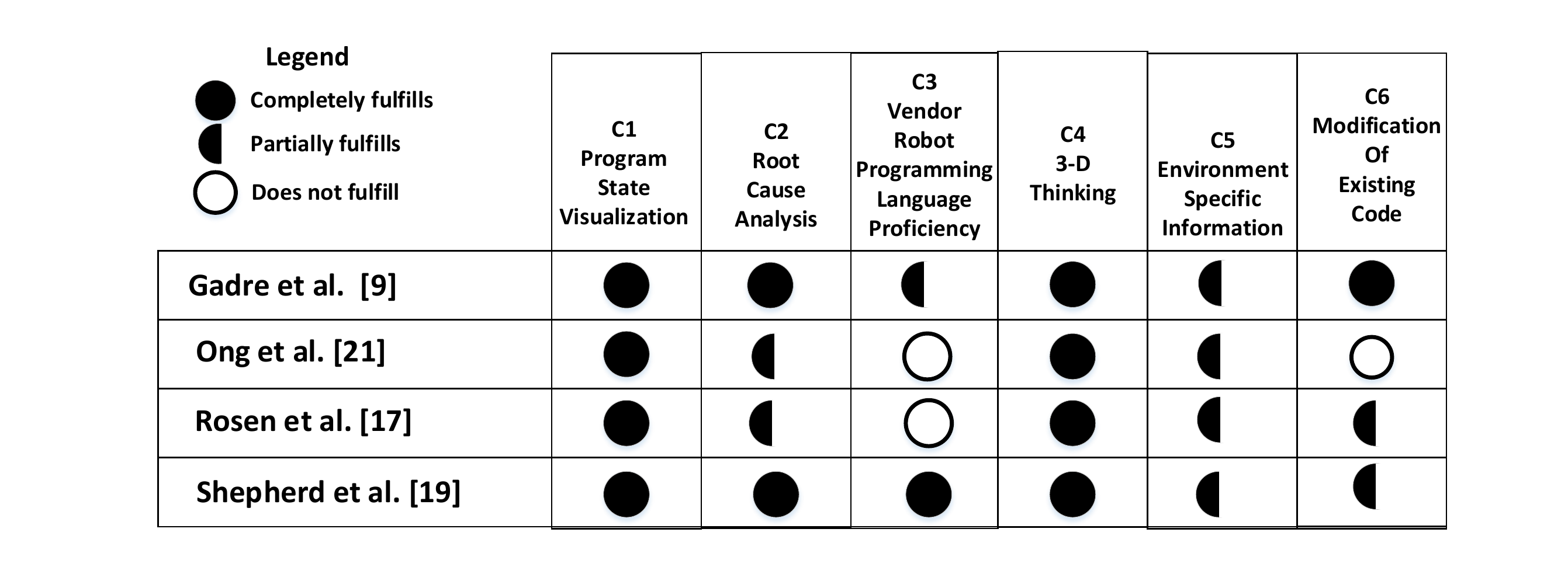}
\caption{Evaluation of AR-based robot programming approaches}
\label{fig:relWorkAR}
\end{figure}

As summarized in Fig.~\ref{fig:relWorkAR}, none of the existing AR-based approaches for robot programming fully address the introduced challenges \textit{C1} - \textit{C6}. Most of these approaches represent robot programs with waypoints, code, or occluding robot arm movements. These representations address \textit{C1}. Hand interaction or pointer interaction, with the User Interface (UI) or with the real robot, as well as a Head-Mounted AR device ensure depth information and human-like interaction with AR objects, thus tackling \textit{C4}. This reflects the identified benefits of using AR. Furthermore, \textit{C3} is either tackled implicitly, by using a high-level block-based Domain-Specific Language (DSL) or not at all, by using no code as a representation. Still, challenge \textit{C5} is not sufficiently covered as integration of environment-specific information is missing e.g. picking up real objects by making a virtual copy. In summary, while identifying and tackling similar challenges as described in this work, the reviewed approaches for AR-based robot programming do not fully support all of them. 

\section{Conceptual Solution}\label{sec:conceptual_solution}

\begin{figure}[hbt!]
\centering
\includegraphics[width=0.85\textwidth]{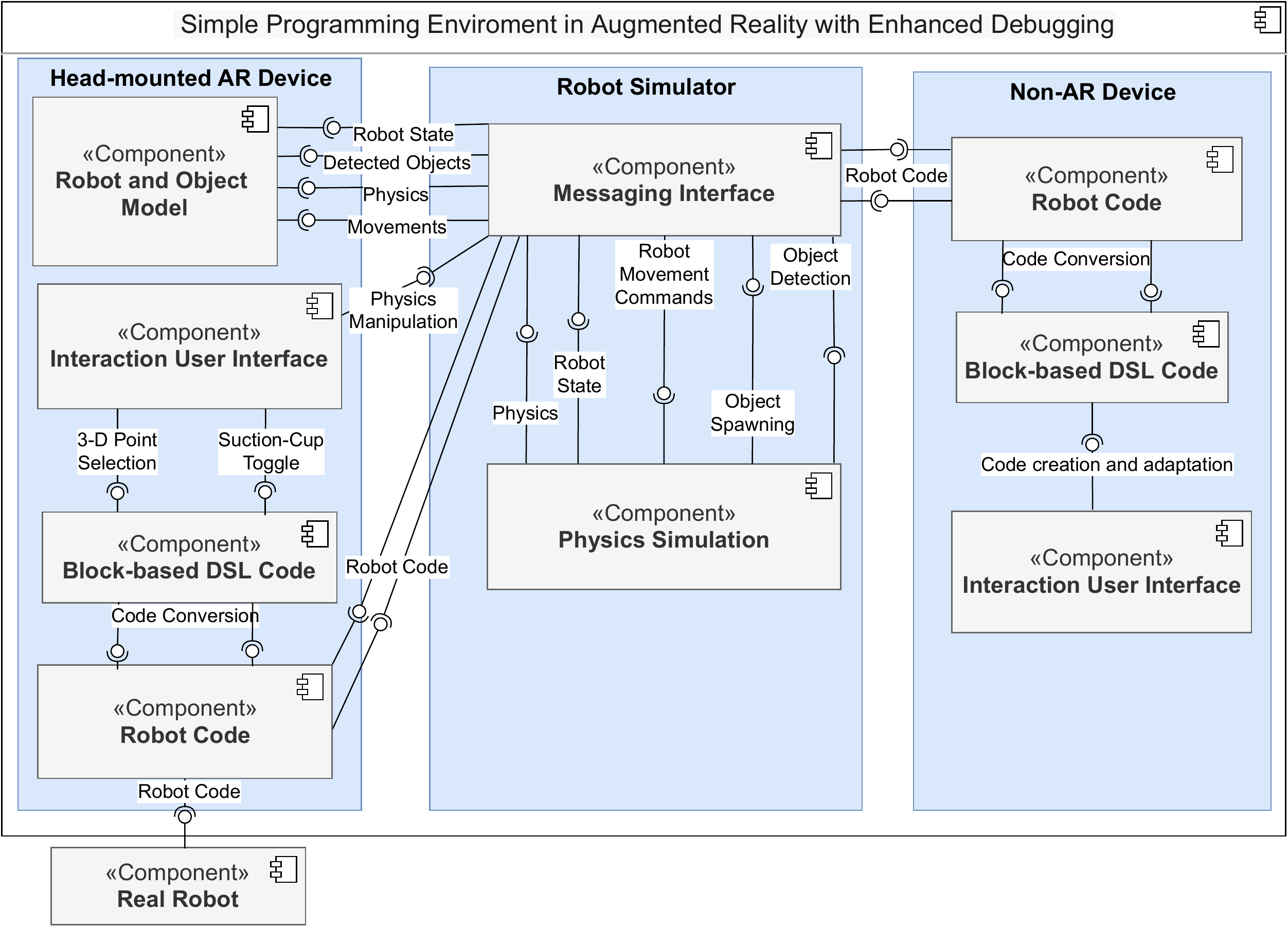}
\caption{An architectural overview of the SPEARED framework}
\label{fig:arch_overview}
\end{figure}

In this section, the conceptual solution of the Simple Programming Environment in Augmented
Reality with Enhanced Debugging (SPEARED) framework is presented. As depicted in Fig.~\ref{fig:arch_overview}, the SPEARED framework can be divided into three main parts: \textit{Head-mounted AR Device}, \textit{Robot Simulator}, and \textit{Non-AR Device}. The \textit{Robot Simulator} is responsible for simulating the robot, its movements, and detected objects. As SPEARED supports robot programming on AR Devices (e.g., HoloLens) and Non-AR Devices (e.g., Laptop or Desktop-PC using an editor), we have two separate components for visualization and interaction: \textit{Head-mounted AR Device} and \textit{Non-AR Device}. In the following, each of the three main parts will be described in more detail.

\subsection{Robot Simulator}\label{subsec:robotSimulator}
As stated in \textit{Challenge C1}, it is necessary to visualize the current program state. The \textit{Robot Simulator} (see Fig.~\ref{fig:arch_overview}) enables the simulated execution of the current program state. Furthermore, it removes the need of setting up a real robot. Please not that if needed, it is possible to register the coordinate system of the virtual robot and the real one. Using a simulation also supports to address \textit{Challenge C2}. Additionally, the \textit{Robot Simulator} enables the simulation of the effect of environmental constraints, as described in \textit{Challenge C5}. The simulation itself depends on two components: The \textit{Physics Simulation} is responsible for realistic robot movements and collisions. The \textit{Messaging Interface} allows communication from and to the simulator. For example, movement commands can be received and executed or the joint states published for other programs. In the following, the \textit{Physics Simulation} and \textit{Messaging Interface} components are described in more detail.


\textbf{Physics Simulation.} The \textit{Physics Simulation} is mimicking real-world physical constraints regarding movements, gravity, etc. It serves as an endpoint for doing robot physics simulation. It enables a simulation of a robot according to predefined properties (e.g., joint position, Computer-Aided Design (CAD) model, etc.) and the execution of movement commands in the same manner as the real robot (e.g. by using inverse kinematics). Additionally, reachability constraints are preserved. Thus, the simulation tackles the execution part of \textit{Challenge C1}. Furthermore, the creation of other objects is possible (e.g., import of existing 3D models). Collisions with these objects are represented realistically. The collision detection is necessary for \textit{Challenge C5}. Finally, the possibility to execute code together with collisions tackles \textit{Challenge C2}. Interfaces to simulation information, e.g., state of the robots’ joints and path trajectories, are available to be consumed. Finally, physical properties, e.g., speed up execution, slow down execution, or change gravity can be manipulated.

\textbf{Messaging Interface.}
The \textit{Messaging Interface} serves as a communication facility for interactions with the \textit{Physics Simulation}. It abstracts from simulation specific endpoints and offers more generic communication methods. Therefore, conversion routines between the requests are necessary. The interface provides endpoints for physics manipulation, robot state, robot movement, and robot code. For physics manipulation, the endpoints speed up, slow down, pause, and resume physics are available to be consumed. For robot state endpoints, the joint states, the robot parts’ orientation and position, the end-effector state, and the idle state can be requested. For robot movement endpoints, the movement command to a specific point in the robot or simulation space can be issued. Furthermore, path trajectories for the planned path can be requested. For robot code endpoints, the interface offers the possibility to convert movement commands to actual robot (simulation) movements. Thus, the \textit{Messaging Interface} offers the endpoints to transfer the execution simulation information for challenges \textit{C1}, \textit{C2}, and \textit{C5} to their respective visualizations.

\subsection{Head-mounted AR device}\label{subsec:arSol}

The AR visualization contains four different components. The \textit{Robot and Object Model}, the \textit{Robot Code}, the \textit{Block-based DSL Code}, and the \textit{Interaction User Interface}. In the following, the components are described and the visualization and interaction possibilities are presented. As described above, SPEARED relies on a robot simulation to visualize and validate created robot programs. To enable users to see these validations in real-time, the simulated robot has to be visualized. The \textit{Robot and Object Model} describes the robot, as well as other detected objects, and its properties e.g. joints, CAD-model, etc. It acts as a representation of the real robot, and detected objects. It is updated based on the movements done in the \textit{Robot Simulator}. Thus, it is necessary for tackling the challenges \textit{C1}, \textit{C2}, and \textit{C5}. The model does not depend on the other components, as it is for visualization purposes only. However, the movement and model information is received from the \textit{Robot Simulator} and not generated inside the AR device itself. The current robot program is displayed in the \textit{Robot Code} component. It allows a vendor-independent representation of e.g. movement and end-effector commands. Thus, it is tackling \textit{Challenge C3}. The programming component of the AR visualization is based on the \textit{Block-based DSL Code} component. It acts as a programming environment. Here, it represents the different movement commands and end-effector commands and enables their adaptation. Thus, it addresses \textit{Challenge C3}.
Furthermore, both the \textit{Robot Code} component as well as the \textit{Block-based DSL Code} component help tackling the challenges \textit{C1} and \textit{C2}.
The \textit{Interaction User Interface} provides interaction and visualization facilities to the user. It visualizes the current robot movement target and the robot end-effector state. The interface enables the user to perform coding activities. These are: Adding new commands, deleting existing commands, modifying existing commands, changing the order of commands, executing the current program, and synchronizing the current program with the non-AR environment. Furthermore, \textit{User Interface (UI)} manipulation is possible. The robot model as well as other UI elements can be moved, rotated, and resized. These operations are specific to the target platform and are supported based on hand gestures on the \textit{Head-mounted AR device} and respective interactions on the \textit{Non-AR Device}. The interface also provides access to the \textit{Robot Simulator} endpoint e.g. the manipulation of physics. Finally, the interaction methods are either gaze, together with hand gestures, or voice commands. The interface enables the challenges \textit{C4} and \textit{C6}.

\subsection{Non-AR device}\label{subsec:non-arSol}

The non-AR interface consists of the components \textit{Robot Code}, \textit{Block-based DSL Code}, and the
\textit{Interaction User Interface}. Its visualization consists of the coding environment, the movement commands, and the end-effector commands. The interface enables the user to perform coding activities. These consist of: adding new commands, deleting existing commands, modifying existing commands, changing the order of commands, and synchronizing the current robot program with the AR environment. Thus, it tackles the challenges \textit{C3} and \textit{C6}.

\subsection{Execution logic and interplay between components}\label{subsec:exLog}

To describe the execution logic of SPEARED in more detail, in the following the intra-component and inter-component interfacing tasks are explained.

\textbf{Intra-Component execution logic.}
Inside the components \textit{Head-mounted AR Device}, \textit{Robot Simulator}, and \textit{Non-AR Device} the  following interfacing tasks exist. In the \textit{Head-Mounted AR Device}, the \textit{Interaction User Interface} allows the parameterization of robot code commands. These parameterizations are shown in Fig.~\ref{fig:arch_overview} on the left side with the labels "3D-Point Selection" and "Suction-Cup Toggle". Therefore, the target of movement commands can be set and the state of the end-effector toggled. Thus, enabling \textit{Challenge C6}. "3-D Point Selection", as in the \textit{Challenge C4}, is the process of setting a movement target by hand and gaze interaction in the AR environment. "Suction-Cup Toggle" is the process of toggling the state of the end-effector e.g. a suction cup. Furthermore, a code-conversion between the created code (with the aforementioned interactions) and robot code is possible. Therefore, the UI elements have to be converted to robot code and vice-versa. This tackles the \textit{Challenge C3}. Inside the \textit{Robot Simulator}, the \textit{Messaging Interface} provides translation capabilities of robot code to movement commands. Here, the robot code has to be translated to physical movements e.g. by using inverse kinematics. Furthermore, the information regarding detected objects, physics, and the robots’ state (e.g. end-effector state) is transferred between the components. This is necessary to allow a realistic simulation and to tackle the challenges \textit{C1}, \textit{C2}, and \textit{C5}. Inside the non-AR device, the Interaction User Interface allows the code creation and adaptation. As a result, the block-based Domain Specific Language (DSL) represents the robot program created. Furthermore, the conversion between the block-based DSL and the robot code is possible. The blocks need to be translated to robot code and vice-versa. This tackles the challenges \textit{C3} and \textit{C6}.

\textbf{Inter-Component execution logic.}
The different devices allow the synchronization of robot code with each other. Therefore, the \textit{Robot Simulator}, or to be specific, the \textit{Messaging Interface}, provides a service to store and load the current robot program. Both devices can upload and download their current code to that place. This also enables \textit{Challenge C6}. Between the \textit{Robot Simulator} and the \textit{Head-mounted AR Device} the robot’s movements, the robot’s state (e.g. end-effector state or whether the robot currently idle), and physics properties are published. This interaction allows the AR device to present the current state of the robot. Furthermore, the detected objects are forwarded to the AR device. This is necessary for the challenges \textit{C1}, \textit{C2}, and \textit{C5}. The manipulation of physics is forwarded to the \textit{Messaging Interface}. Implicitly, the conversion between coordinate systems of the AR-device and \textit{Robot Simulator} has to be done, when detected objects or movements are transferred. This is required for the \textit{Challenge C4}. Additionally, the code can be executed on the \textit{Real Robot}. 

\section{Implementation}\label{sec:implementation}
\begin{figure}[hbt!]
\centering
\includegraphics[width=0.85\textwidth]{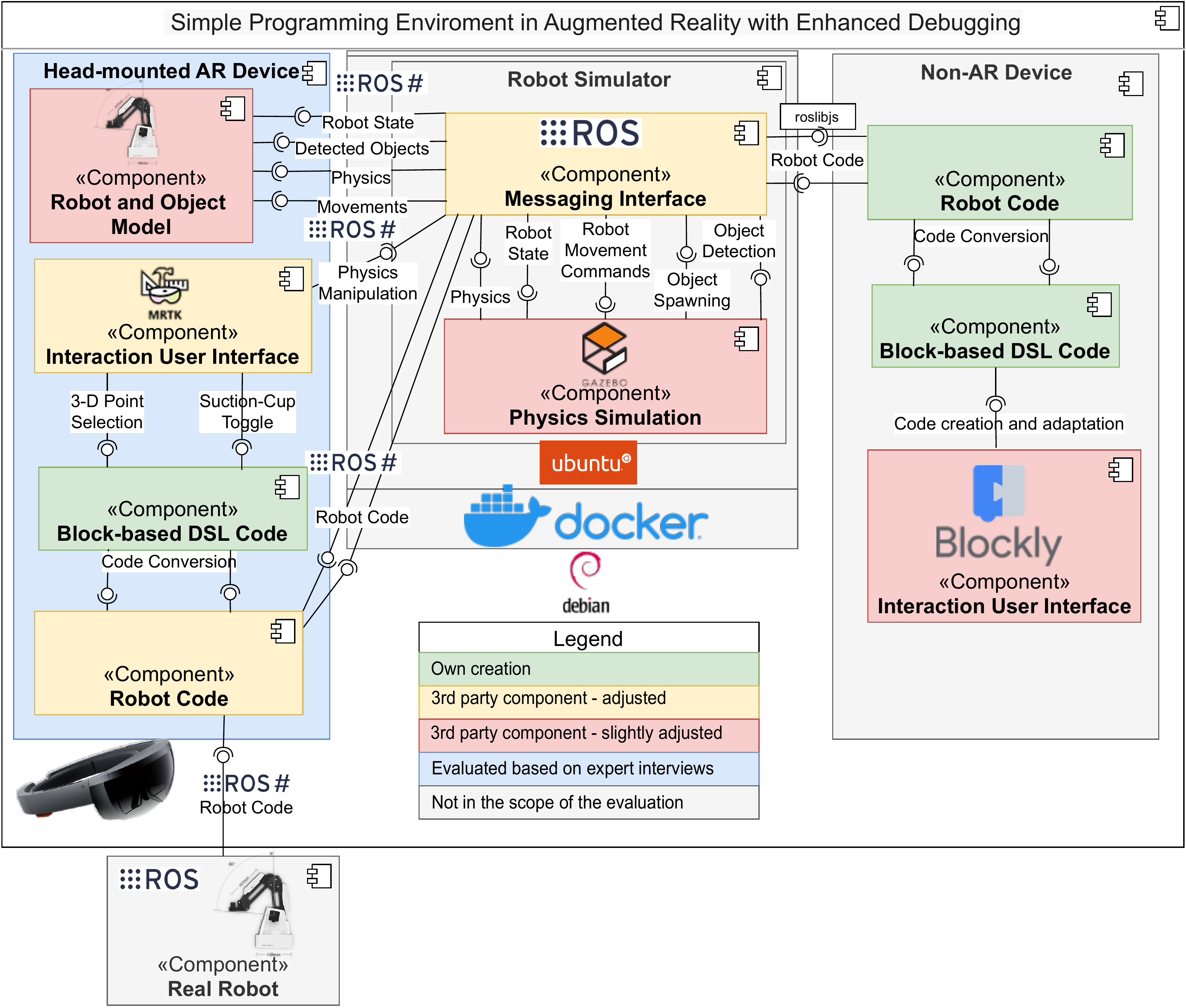}
\caption{Used technologies in SPEARED}
\label{fig:imp_overview}
\end{figure}

In this section, we describe implementation-specific details of the SPEARED framework which is publicly available as an open-source software project at GitHub\footnote{https://github.com/VARobot-PG/application}. 
Fig. \ref{fig:imp_overview} shows the architecture of the SPEARED framework annotated with the used technologies. 

\newpage
\textbf{Robot Simulator}

The \textit{Robot Simulator} is based on \textit{Gazebo\footnote{http://gazebosim.org}} as a \textit{Physics Simulation} and \textit{Robot Operating System (ROS)\footnote{https://www.ros.org}} as a \textit{Messaging Interface}. It is deployed in a \textit{Docker\footnote{https://www.docker.com}} container using Ubuntu.
To mimic the behavior of the real robot, a \textit{Physics Simulation} is needed. For this purpose, \textit{Gazebo} was chosen which is able to simulate robots and physics as well as offers graphical and programmatic interfaces. To simulate the robot in \textit{Gazebo}, the robot model has to be created. It is based on an XML macro (XACRO) file \footnote{https://ni.www.techfak.uni-bielefeld.de/files/URDF-XACRO.pdf}. It is spawned inside the simulation context. This robot model describes the different properties of the robot. It includes the visuals, the physics e.g. mass, etc., and the collisions for each part of the robot. Additionally, \textit{Gazebo} offers ROS endpoints for physics manipulation. The \textit{Messaging Interface} component is based on ROS, which provides a communication interface based on asynchronous and synchronous communication. ROS offers a parameter server that enables the sharing of relevant information. Here, the server is used for code synchronization. Furthermore, topics are used for publishing the current position of the robot arm or information about whether the robot is idle. The Real Robot is a \textit{DoBot Magician\footnote{https://www.dobot.us/}}. It offers the same ROS endpoints as the AR simulation. Hence, on the implementation side, it does not matter whether the code is being sent to the real robot or the simulation. 

\textbf{Head-mounted AR Device}

\begin{figure}[hbt!]
\centering
\includegraphics[width=0.8\textwidth]{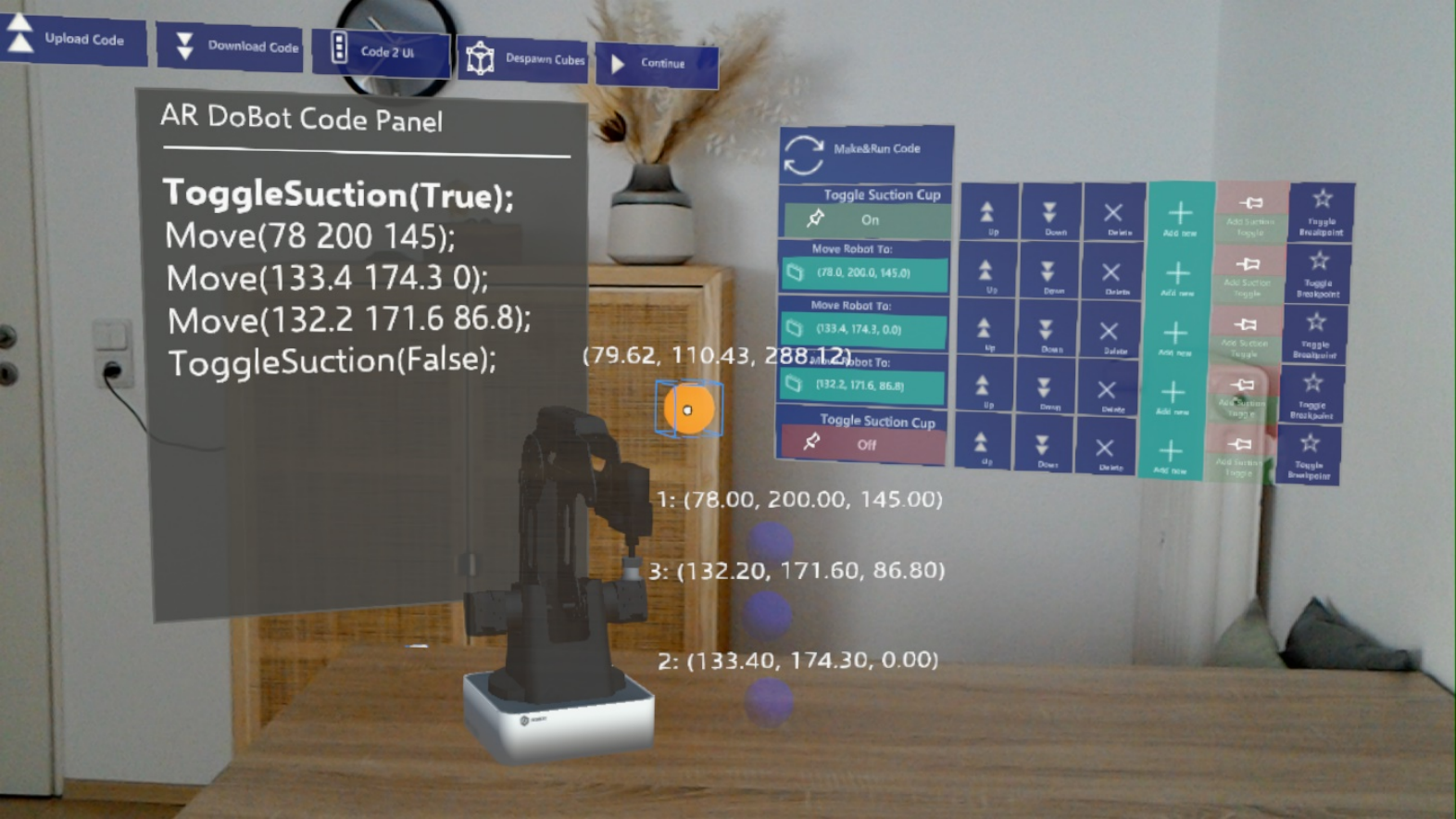}
\caption{Robot Code, Robot Model in an AR application and block-based DSL Code (f.l.t.r.)}
\label{fig:arDobotCodePanel}
\end{figure}

As a \textit{Head-mounted AR-Device}, the Microsoft HoloLens was chosen, while the Unity game engine together with the Microsoft MRTK\footnote{https://github.com/microsoft/MixedRealityToolkit-Unity} was used for application development purposes. As mentioned in the previous section, the AR application consists of the components \textit{Robot and Object Model}, \textit{Robot Code}, \textit{Blockbased DSL Code}, and the \textit{Interaction User Interface} (see Fig. \ref{fig:arDobotCodePanel}).  

\begin{figure}[hbt!]
\centering
\includegraphics[width=0.8\textwidth]{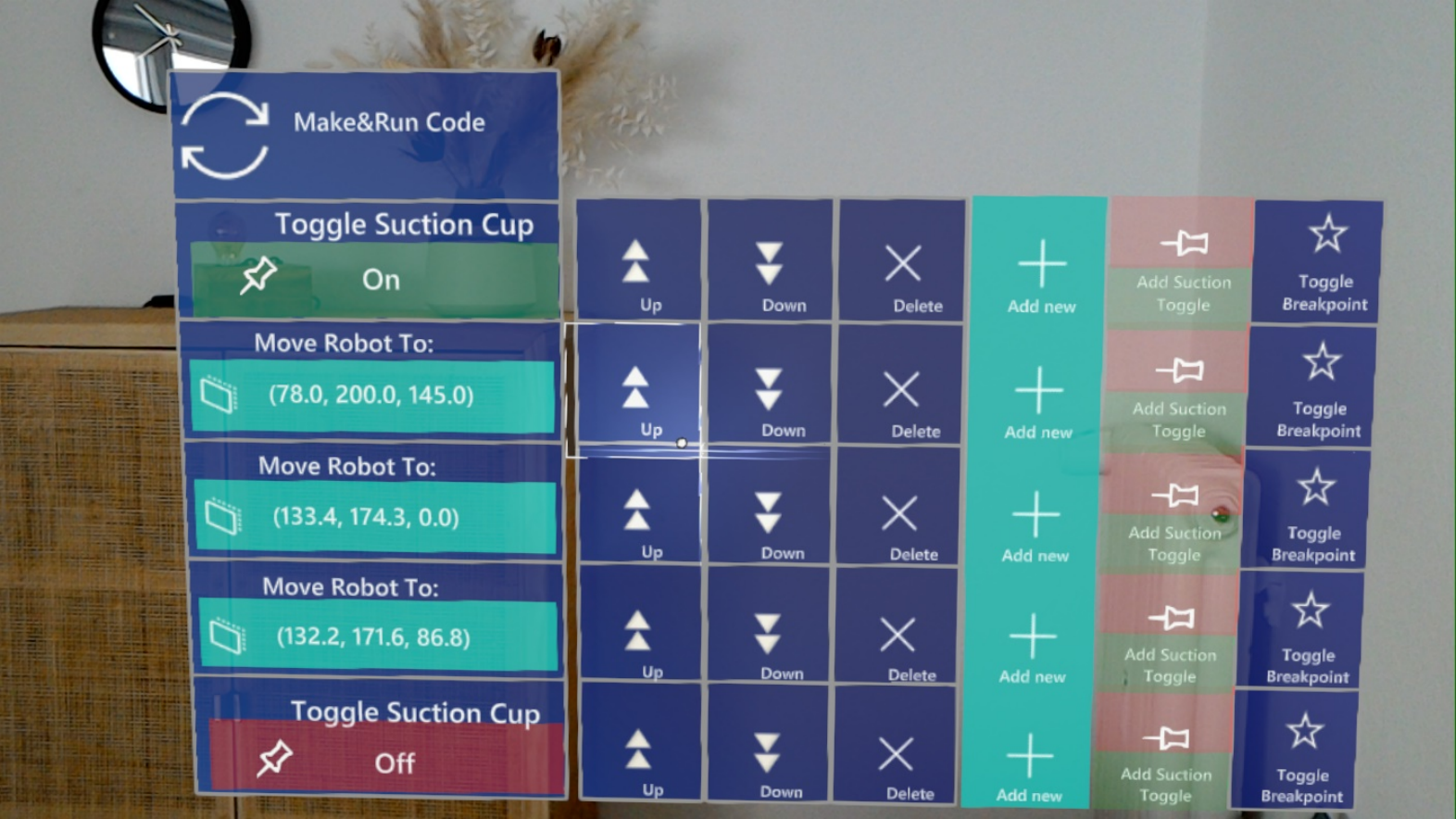}
\caption{AR Block-based coding environment}
\label{fig:screenshot2}
\end{figure}

The \textit{Robot and Object Model} represents the current robot's state. Here, the robot model,
as well as detected objects, are shown. Their visual representation is based on a XACRO file (see above) and a Simulation Description Format (SDF) file respectively. The visualization of detected objects is necessary for \textit{Challenge C5}. Here, the aforementioned self-programmed service is used. This endpoint can be triggered manually. The current orientation of the different robot parts is adapted at runtime. The model reflects the current program execution. It is needed for challenges \textit{C1}, \textit{C2}, and \textit{C5}. The model itself is added as a GameObject in Unity. The \textit{Block-based DSL Code} is controlled via the \textit{Interaction User Interface}. On the programming side, commands can be created, deleted, their order can be changed, and the commands’ parameter can be set e.g. switch end-effector on, switch end-effector off, and set target coordinate for movement commands. For selecting the target position of the movement command, a target sphere is moved to the wanted position (refer to the yellow sphere in Fig. \ref{fig:arDobotCodePanel}), which enables both the definition and visualization for \textit{Challenge C4}. The current coordinate of the sphere can be selected as a target by pressing the respective move-statement in the \textit{Block-based DSL Code} component (Fig.~\ref{fig:screenshot2}).

The \textit{Block-based DSL Code} component, shown in Fig.~\ref{fig:screenshot2}, together with its interaction capabilities via the \textit{Interaction User Interface} component, is needed for the challenges \textit{C6}, \textit{C3}, \textit{C4} for coordinate definition purposes.

\textbf{Non-AR Device.}

The \textit{Non-AR device} solution is based on \textit{Google Blockly\footnote{https://developers.google.com/blockly}} and \textit{roslibjs\footnote{https://github.com/RobotWebTools/roslibjs}} for interfacing purposes.
It consists of the \textit{Interaction User Interface}, the \textit{Block-based DSL}, and the \textit{Robot Code} component. It resembles the programming experience of a simple Integrated Development Environment (IDE).

\begin{figure}[hbt!]
\centering
\includegraphics[width=0.8\textwidth]{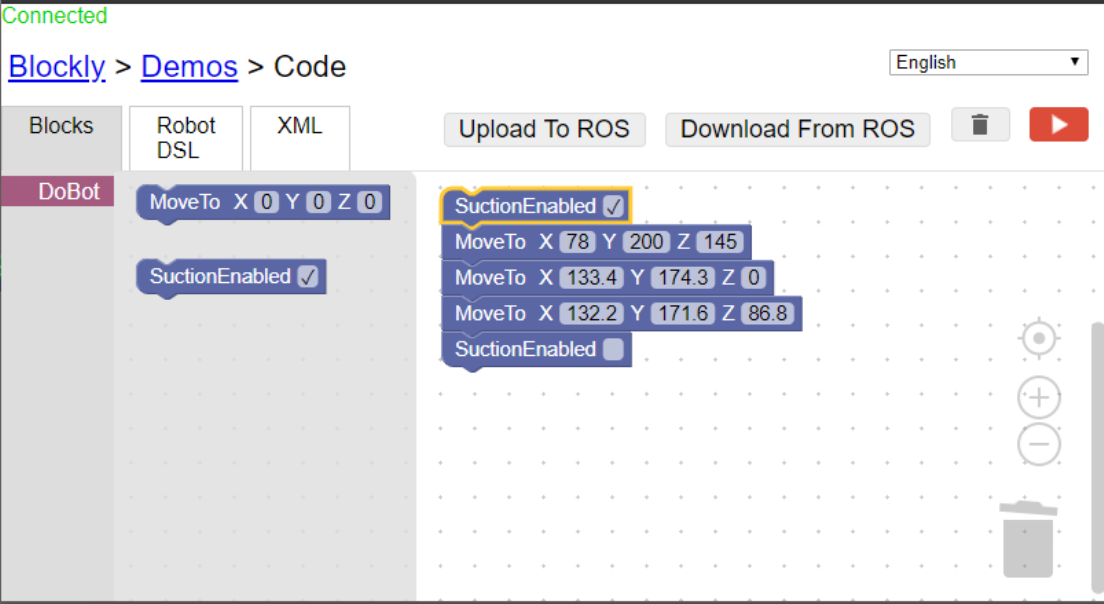}
\caption{Adapted Blockly environment with custom blocks (above) and for robot code (below)}
\label{fig:screenshot4}
\end{figure}

Additionally, code modifications can be synchronized between the non-AR and the AR environment. To provide a uniform experience across the different devices, a similar \textit{Block-based DSL} is used. The \textit{Block-based DSL} is based on custom blocks. They were defined for movement commands, which consist of three coordinates, and end-effector commands, which enable or disable the suction cup. Here, the name as well as e.g. parameters and their types had to be set. This enables the usage of features independent from the vendor language, thus, tackling \textit{Challenge C3}. The adaptation of values tackles \textit{Challenge C6}. The layout with instances of these definitions can be seen in Fig. \ref{fig:screenshot4}.

\section{Evaluation}\label{sec:usability_evaluation}
For the evaluation of the SPEARED framework concerning the introduced challenges in Sec.~\ref{sec:intro} we have conducted expert interviews. In the following, we first describe the setup and execution of the expert interviews. After that, we present and discuss its main results. 

\subsection{Setup and execution of expert interviews}
\label{subsec:usability_setup}

\begin{figure}[ht!]
\centering
\includegraphics[width=\textwidth]{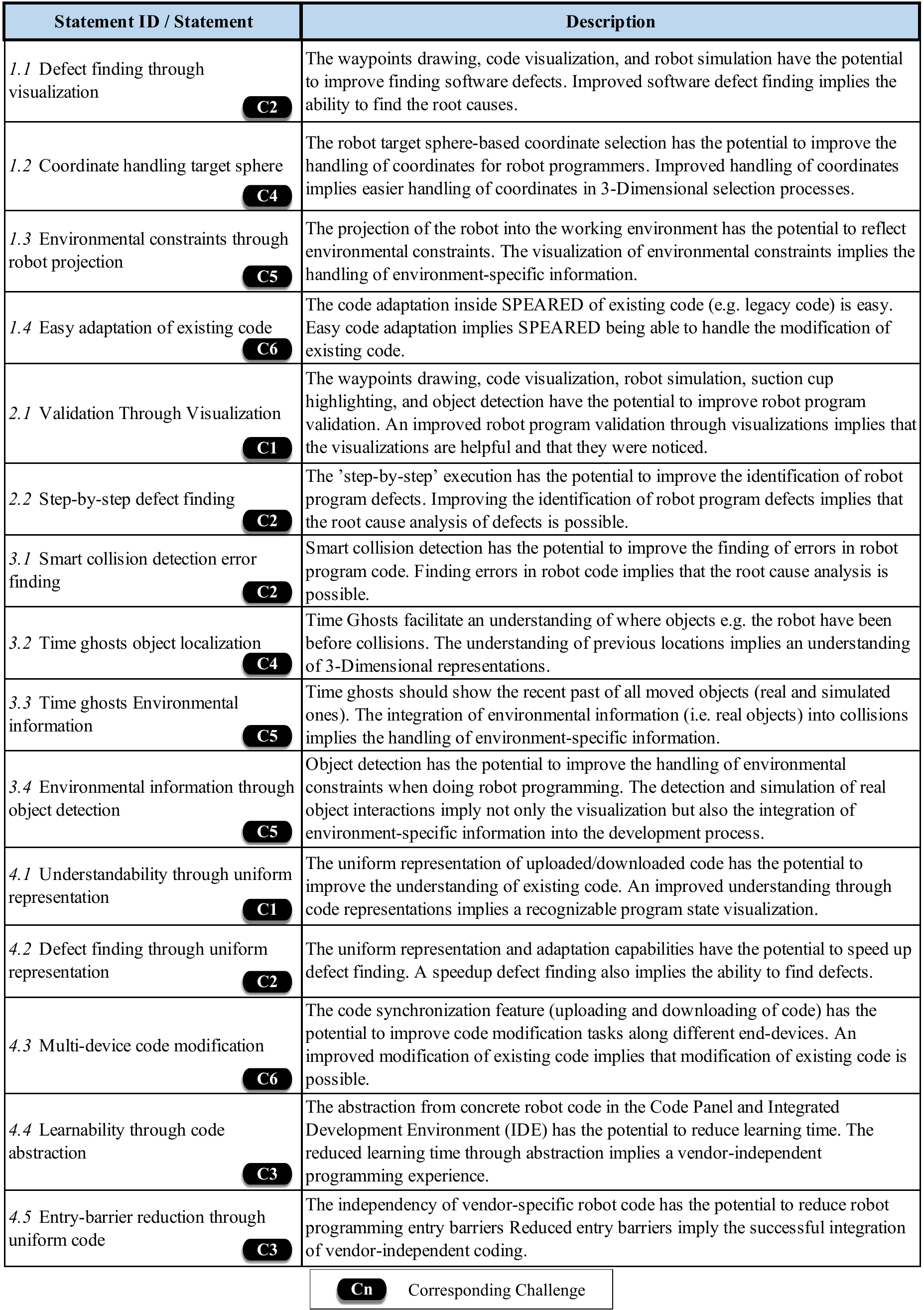}
\caption{Overview of the main statements and their corresponding challenges}
\label{fig:challengesOverview}
\end{figure}

Unfortunately, due to the current situation (COVID-19), we could not conduct usability tests with a larger group of heterogeneous end-users as initially planned. Instead expert interviews were conducted as semi-structured interviews \cite{newcomer2015conducting}. 
We conducted expert interviews with 9 experts from the area of robotics, automation, and augmented reality. Inside the interview, a pre-defined question set was asked to every participant. To derive the questions for the expert interviews, a GQM (Goal Question Metric) model is used \cite{gqm-book}. The evaluation consists of six goals. These goals describe whether SPEARED fulfills the respective challenges motivated in Sec.~\ref{sec:intro}. For each goal, different statements, describing specific components of the respective challenge, were created (see Fig.~\ref{fig:challengesOverview}).

\subsection{Evaluation results}
\label{subsec:eval_results}

\begin{figure}[hbt!]
\centering
\includegraphics[width=\textwidth]{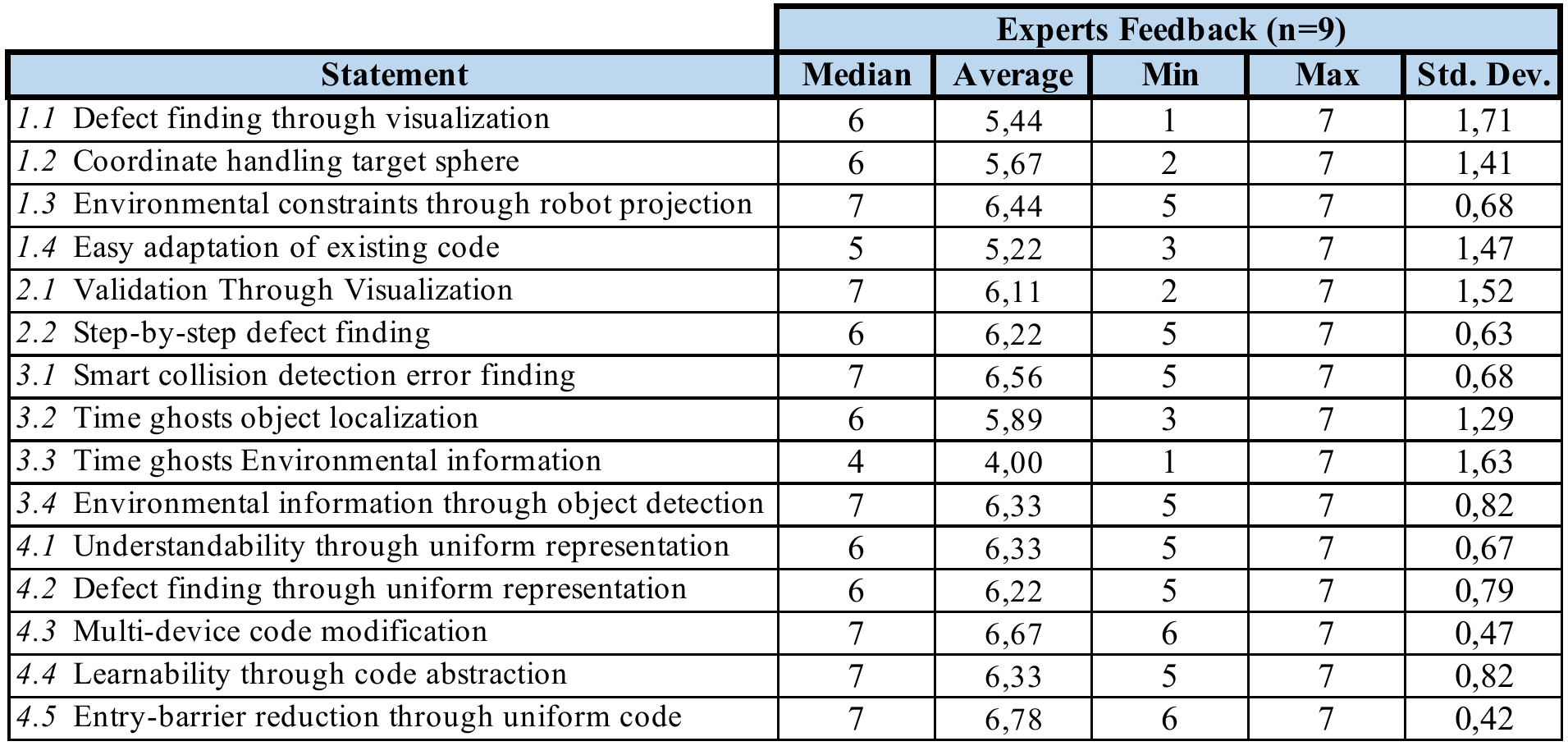}
\caption{Feedback to Statements Overview}
\label{fig:feedbackOverview}
\end{figure}

\begin{figure}[hbt!]
\centering
\includegraphics[width=\textwidth]{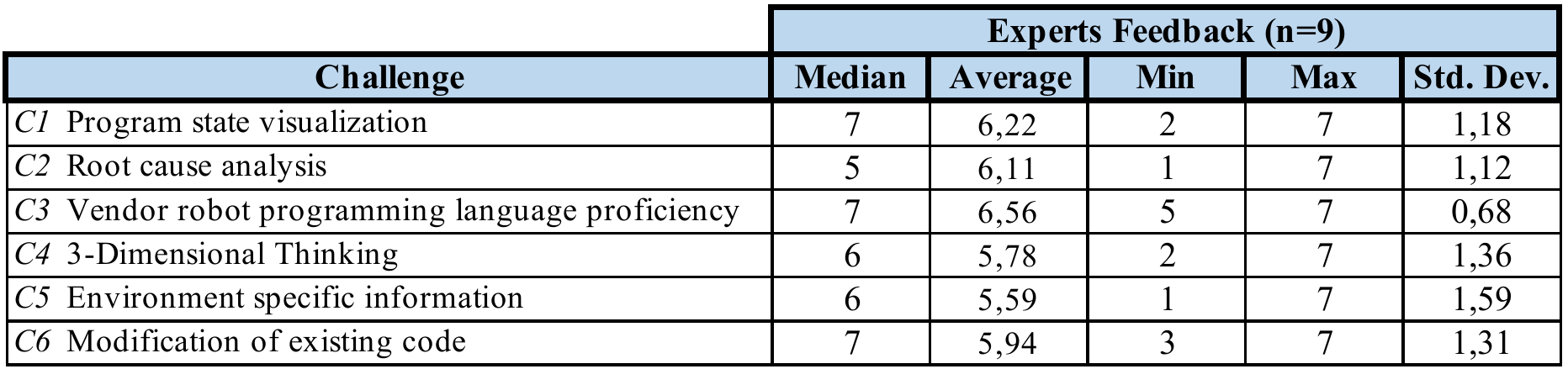}
\caption{Feedback to Challenges Overview}
\label{fig:feedbackChallengesOverview}
\end{figure}

In the following, the answers, expressing the expert’s degree of agreement to the provided
statements are presented. For each statement, the experts’ degree of agreement was measured on a 7-point Likert scale. For evaluation purposes, metrics on a per statement and a per challenge basis were used where the median, arithmetic mean, minimum rating, and standard deviation of the aggregated answers were calculated. An overview of the extracted metrics - per statement - is depicted in Fig.~\ref{fig:feedbackOverview}. 
As explained before, the statements are mapped to the challenges. To evaluate the challenges,
the answers to the respective statements are combined e.g. \textit{Challenge 2} is equal to the
combination of the answers to Statement 1.1., 2.2, 3.1, and 4.2 (Fig.~\ref{fig:challengesOverview}). Every single feedback of each statement mapped to a challenge is aggregated to calculate the median, average (arithmetic mean), minimum, and standard deviation on a challenge level. The results, showing an overview of the aggregated metrics - per challenge - is shown in Fig.~\ref{fig:feedbackChallengesOverview}. 

Based on the results above and the informal feedback we have received, we will discuss in the following how the experts rated the SPEARED framework in addressing the challenges.
 
\textbf{Challenge 1}: The majority of the experts agreed that the AR-based visualization and simulation of the robot helps to increase the understandability of the robot code. Especially the visualization of robot part information, e.g., the suction cup status (on/off) was identified as helpful. Here, one expert concluded that this status shows the potential to integrate other helpful information, e.g., current weight load, heat, and motor force. Still, the experts claimed that the current visualization state is not end-user ready and potentially hard to understand without prior introduction. Nevertheless, they see the potential of the approach and believe that further development and better AR hardware will circumvent the current User Interface (UI) problems. 

\textbf{Challenge 2}: In general, the experts see SPEARED’s potential to aid in root cause analysis processes. Especially the smart collision detection feature using the benefits of AR for object recognition and projection facilities was agreed to have potential to improve defect finding. However, some experts noted the fact that some of the existing solutions already provide sophisticated ways of finding software defects through intelligent debugging mechanisms. With this regard they suggested to enrich those solutions with AR capabilities to combine the strength of both solutions.

\textbf{Challenge 3}: The experts expressed that the abstraction from concrete robot code is especially helpful for lay people as they do not need to learn a new language every time they change the robot. Furthermore, it was stated that the focus on the Block-based Domain Specific Language (DSL) may also reduce the possible mistakes and ensures a smoother programming process for beginners. This goes hand in hand with the feedback to \textit{Statement 4.5} which received two ratings of "6 - Agree" and seven ratings of "7 - Strongly Agree". However, multiple experts stated that they see the advantages mainly in supporting lay people in the programming process and further more sophisticated or fine grained commands could be provided to also support the experts.

\textbf{Challenge 4}: The received feedback suggests that coordinate handling is challenging and SPEARED does help in tackling these difficulties with the coordinate selection mechanism and the waypoints drawing. One of the experts stated that problems may occur when trying to select coordinates where different axial combinations are possible. This feature was intentionally not implemented in SPEARED due to simplicity reasons. Also, SPEARED is using a sample robot system (Dobot Magician) which does not have different axial combinations to reach a specific coordinate. Nevertheless, the experts agreed that SPEARED supports "3-dimensional thinking" which is enabled through the projection of the simulated robot arm in the real-world setting. 

\textbf{Challenge 5}: In general, the experts agreed that SPEARED’s object detection has the potential, to improve the handling of environmental constraints when doing robot programming. This feedback also shows that the experts see the representation and integration of environmental constraints into the programming environment as beneficial. However, Statement 3.3 received one rating of "1- Strongly Disagree" and one "2 - Disagree" with the others being "4 - Neutral" or higher. Here, the feedback of the experts showed that either the meaning of the statement was quite unclear or the experts opinions differed. Nevertheless, the majority of the informal feedback indicates that that it is helpful when environmental context information are integrated into the AR programming environment e.g. pick yellow cube. Hence, the realization of integration of environment specific information has been considered as a helpful feature for AR-assisted robot programming. 

\textbf{Challenge 6}: Concerning the addressing of this challenge, we received different opinions from the experts. This leads to believe that the experts were not sure about the feature or the meaning of the statement. The ambiguity of the statement is also reflected in one expert asking for explicitly stating the meaning of legacy code and its adaptation capabilities in the context of the statement. However, after explaining the meaning, most experts agreed that SPEARED offers easy code adaptation capabilities. All experts were sure that SPEARED has potential to improve code modification tasks along different end-devices. 

The discussion above shows preliminary results regarding the fulfillment of the challenges. However, multiple experts stated that some statements were subjectively unclear for them. Here, the authors tried to explain the respective statement without being too biased. Still, technical
difficulties like video stuttering, audio delays, and missing webcams made the process of remote expert interviews more challenging and may lead to additional bias.

\section{Conclusion and Outlook}\label{sec:conclusion}

The usage of robots promises support in a plethora of activities in e.g. households, education,
care, and businesses. However, robot programming faces high entry barriers due to the complexity of robot programming itself. Even for experts robot programming is a cumbersome and error-prone task. In this paper, an Augmented Reality (AR)-assisted robot programming environment is presented. Usability evaluation results based on expert interviews show that the solution approach has the potential to enrich current robot programming processes and thus could reduce complexity and entry barriers of robot programming.

In future work, we plan to investigate the efficiency, effectiveness, and user satisfaction of the SPEARED framework based on usability tests with end-users from different domains. Furthermore, Programming by Demonstration (PbD) could be used to further simplify the programming complexity of robots. PbD allows users to program robots by demonstrating a task. Using PbD for generating source code in the AR Framework, which can then be adopted, could allow an even more intuitive way of programming. 

\subsubsection*{Acknowledgement} We would like to thank Jonas Eilers and Michael Wieneke for their support during the implementation and evaluation of the presented approach.


\bibliographystyle{splncs04}
\bibliography{references}

\begin{thebibliography}{10}
\providecommand{\url}[1]{\texttt{#1}}
\providecommand{\urlprefix}{URL }
\providecommand{\doi}[1]{https://doi.org/#1}

\bibitem{DBLP:journals/ras/AleottiCR04}
Aleotti, J., Caselli, S., Reggiani, M.: Leveraging on a virtual environment for
  robot programming by demonstration. Robotics Auton. Syst.  \textbf{47}(2-3),
  153--161 (2004). \doi{10.1016/j.robot.2004.03.009},
  \url{https://doi.org/10.1016/j.robot.2004.03.009}

\bibitem{DBLP:journals/tsmc/AleottiCR05}
Aleotti, J., Caselli, S., Reggiani, M.: Evaluation of virtual fixtures for a
  robot programming by demonstration interface. {IEEE} Trans. Syst. Man Cybern.
  Part {A}  \textbf{35}(4),  536--545 (2005). \doi{10.1109/TSMCA.2005.850604},
  \url{https://doi.org/10.1109/TSMCA.2005.850604}

\bibitem{DBLP:conf/have/AleottiMC14}
Aleotti, J., Micconi, G., Caselli, S.: Programming manipulation tasks by
  demonstration in visuo-haptic augmented reality. In: 2014 {IEEE}
  International Symposium on Haptic, Audio and Visual Environments and Games,
  {HAVE} 2014, Richardson, TX, USA, October 10-11, 2014. pp. 13--18. {IEEE}
  (2014). \doi{10.1109/HAVE.2014.6954324},
  \url{https://doi.org/10.1109/HAVE.2014.6954324}

\bibitem{DBLP:conf/rss/AlexandrovaCHT14}
Alexandrova, S., Cakmak, M., Hsiao, K., Takayama, L.: Robot programming by
  demonstration with interactive action visualizations. In: Fox, D., Kavraki,
  L.E., Kurniawati, H. (eds.) Robotics: Science and Systems X, University of
  California, Berkeley, USA, July 12-16, 2014 (2014).
  \doi{10.15607/RSS.2014.X.048},
  \url{http://www.roboticsproceedings.org/rss10/p48.html}

\bibitem{DBLP:journals/vlc/CoronadoMIV20}
Coronado, E., Mastrogiovanni, F., Indurkhya, B., Venture, G.: Visual
  programming environments for end-user development of intelligent and social
  robots, a systematic review. J. Comput. Lang.  \textbf{58},  100970 (2020).
  \doi{10.1016/j.cola.2020.100970},
  \url{https://doi.org/10.1016/j.cola.2020.100970}

\bibitem{Fang2012}
Fang, H., Ong, S.K., Nee, A.: Orientation planning of robot end-effector using
  augmented reality. The International Journal of Advanced Manufacturing
  Technology  \textbf{67} (08 2012). \doi{10.1007/s00170-012-4629-7}

\bibitem{Fang2014}
Fang, H., Ong, S.K., Nee, A.: Novel ar-based interface for human-robot
  interaction and visualization. Advances in Manufacturing  \textbf{2} (12
  2014). \doi{10.1007/s40436-014-0087-9}

\bibitem{DBLP:conf/vw/FangON09}
Fang, H., Ong, S., Nee, A.Y.C.: Robot programming using augmented reality. In:
  Ugail, H., Qahwaji, R., Earnshaw, R.A., Willis, P.J. (eds.) 2009
  International Conference on CyberWorlds, Bradford, West Yorkshire, UK, 7-11
  September 2009. pp. 13--20. {IEEE} Computer Society (2009).
  \doi{10.1109/CW.2009.14}, \url{https://doi.org/10.1109/CW.2009.14}

\bibitem{DBLP:conf/icra/GadreRCPTK19}
Gadre, S.Y., Rosen, E., Chien, G., Phillips, E., Tellex, S., Konidaris, G.D.:
  End-user robot programming using mixed reality. In: International Conference
  on Robotics and Automation, {ICRA} 2019, Montreal, QC, Canada, May 20-24,
  2019. pp. 2707--2713. {IEEE} (2019). \doi{10.1109/ICRA.2019.8793988},
  \url{https://doi.org/10.1109/ICRA.2019.8793988}

\bibitem{DBLP:conf/taros/GianniFP13}
Gianni, M., Ferri, F., Pirri, F.: {ARE:} augmented reality environment for
  mobile robots. In: Natraj, A., Cameron, S., Melhuish, C., Witkowski, M.
  (eds.) Towards Autonomous Robotic Systems - 14th Annual Conference, {TAROS}
  2013, Oxford, UK, August 28-30, 2013, Revised Selected Papers. Lecture Notes
  in Computer Science, vol.~8069, pp. 470--483. Springer (2013).
  \doi{10.1007/978-3-662-43645-5\_48},
  \url{https://doi.org/10.1007/978-3-662-43645-5\_48}

\bibitem{DBLP:conf/hcse/GottschalkYSE20}
Gottschalk, S., Yigitbas, E., Schmidt, E., Engels, G.: Model-based product
  configuration in augmented reality applications. In: Bernhaupt, R., Ardito,
  C., Sauer, S. (eds.) Human-Centered Software Engineering - 8th {IFIP} {WG}
  13.2 International Working Conference, {HCSE} 2020, Eindhoven, The
  Netherlands, November 30 - December 2, 2020, Proceedings. Lecture Notes in
  Computer Science, vol. 12481, pp. 84--104. Springer (2020).
  \doi{10.1007/978-3-030-64266-2\_5},
  \url{https://doi.org/10.1007/978-3-030-64266-2\_5}

\bibitem{DBLP:conf/hcse/GottschalkYSE20a}
Gottschalk, S., Yigitbas, E., Schmidt, E., Engels, G.: Proconar: {A} tool
  support for model-based {AR} product configuration. In: Bernhaupt, R.,
  Ardito, C., Sauer, S. (eds.) Human-Centered Software Engineering - 8th {IFIP}
  {WG} 13.2 International Working Conference, {HCSE} 2020, Eindhoven, The
  Netherlands, November 30 - December 2, 2020, Proceedings. Lecture Notes in
  Computer Science, vol. 12481, pp. 207--215. Springer (2020).
  \doi{10.1007/978-3-030-64266-2\_14},
  \url{https://doi.org/10.1007/978-3-030-64266-2\_14}

\bibitem{DBLP:conf/hcse/JovanovikjY0E20}
Jovanovikj, I., Yigitbas, E., Sauer, S., Engels, G.: Augmented and virtual
  reality object repository for rapid prototyping. In: Bernhaupt, R., Ardito,
  C., Sauer, S. (eds.) Human-Centered Software Engineering - 8th {IFIP} {WG}
  13.2 International Working Conference, {HCSE} 2020, Eindhoven, The
  Netherlands, November 30 - December 2, 2020, Proceedings. Lecture Notes in
  Computer Science, vol. 12481, pp. 216--224. Springer (2020).
  \doi{10.1007/978-3-030-64266-2\_15},
  \url{https://doi.org/10.1007/978-3-030-64266-2\_15}

\bibitem{DBLP:conf/eics/KringsYJ0E20}
Krings, S., Yigitbas, E., Jovanovikj, I., Sauer, S., Engels, G.: Development
  framework for context-aware augmented reality applications. In: Bowen, J.,
  Vanderdonckt, J., Winckler, M. (eds.) {EICS} '20: {ACM} {SIGCHI} Symposium on
  Engineering Interactive Computing Systems, Sophia Antipolis, France, June
  23-26, 2020. pp. 9:1--9:6. {ACM} (2020). \doi{10.1145/3393672.3398640},
  \url{https://doi.org/10.1145/3393672.3398640}

\bibitem{Lambrecht2011}
{Lambrecht}, J., {Chemnitz}, M., {Krüger}, J.: Control layer for multi-vendor
  industrial robot interaction providing integration of supervisory process
  control and multifunctional control units. In: 2011 IEEE Conference on
  Technologies for Practical Robot Applications. pp. 115--120 (2011).
  \doi{10.1109/TEPRA.2011.5753492}

\bibitem{DBLP:journals/bise/LasiFKFH14}
Lasi, H., Fettke, P., Kemper, H., Feld, T., Hoffmann, M.: Industry 4.0. Bus.
  Inf. Syst. Eng.  \textbf{6}(4),  239--242 (2014).
  \doi{10.1007/s12599-014-0334-4},
  \url{https://doi.org/10.1007/s12599-014-0334-4}

\bibitem{newcomer2015conducting}
Newcomer, K.E., Hatry, H.P., Wholey, J.S.: Conducting semi-structured
  interviews. Handbook of practical program evaluation  \textbf{492} (2015)

\bibitem{7745110}
{Orendt}, E.M., {Fichtner}, M., {Henrich}, D.: Robot programming by
  non-experts: Intuitiveness and robustness of one-shot robot programming. In:
  2016 25th IEEE International Symposium on Robot and Human Interactive
  Communication (RO-MAN). pp. 192--199 (2016). \doi{10.1109/ROMAN.2016.7745110}

\bibitem{DBLP:conf/isr/PanPLDN10}
Pan, Z., Polden, J., Larkin, N., van Duin, S., Norrish, J.: Recent progress on
  programming methods for industrial robots. In: {ISR/ROBOTIK} 2010,
  Proceedings for the joint conference of {ISR} 2010 (41st Internationel
  Symposium on Robotics) und {ROBOTIK} 2010 (6th German Conference on
  Robotics), 7-9 June 2010, Munich, Germany - Parallel to {AUTOMATICA}.
  pp.~1--8. {VDE} Verlag (2010),
  \url{http://ieeexplore.ieee.org/document/5756855/}

\bibitem{DBLP:conf/ismar/PettersenPSEL03}
Pettersen, T., Pretlove, J., Skourup, C., Engedal, T., L{\o}kstad, T.:
  Augmented reality for programming industrial robots. In: 2003 {IEEE} / {ACM}
  International Symposium on Mixed and Augmented Reality {(ISMAR} 2003), 7-10
  October 2003, Tokyo, Japan. pp. 319--320. {IEEE} Computer Society (2003).
  \doi{10.1109/ISMAR.2003.1240739},
  \url{https://doi.org/10.1109/ISMAR.2003.1240739}

\bibitem{DBLP:journals/ijrr/RosenWPCTKT19}
Rosen, E., Whitney, D., Phillips, E., Chien, G., Tompkin, J., Konidaris, G.D.,
  Tellex, S.: Communicating and controlling robot arm motion intent through
  mixed-reality head-mounted displays. Int. J. Robotics Res.
  \textbf{38}(12-13) (2019). \doi{10.1177/0278364919842925},
  \url{https://doi.org/10.1177/0278364919842925}

\bibitem{DBLP:conf/scam/ShepherdFWFLA18}
Shepherd, D.C., Francis, P., Weintrop, D., Franklin, D., Li, B., Afzal, A.:
  [engineering paper] an {IDE} for easy programming of simple robotics tasks.
  In: 18th {IEEE} International Working Conference on Source Code Analysis and
  Manipulation, {SCAM} 2018, Madrid, Spain, September 23-24, 2018. pp.
  209--214. {IEEE} Computer Society (2018). \doi{10.1109/SCAM.2018.00032},
  \url{https://doi.org/10.1109/SCAM.2018.00032}

\bibitem{DBLP:conf/icse/ShepherdKF19}
Shepherd, D.C., Kraft, N.A., Francis, P.: Visualizing the "hidden" variables in
  robot programs. In: Proceedings of the 2nd International Workshop on Robotics
  Software Engineering, RoSE 2019, Montreal, QC, Canada, May 27, 2019. pp.
  13--16. {IEEE} / {ACM} (2019). \doi{10.1109/RoSE.2019.00007},
  \url{https://doi.org/10.1109/RoSE.2019.00007}

\bibitem{ong}
Thanigaivel, N.K., Ong, S.K., Nee, A.: Augmented reality-assisted robot
  programming system for industrial applications. Robotics and
  Computer-Integrated Manufacturing  \textbf{61} (06 2019).
  \doi{10.1016/j.rcim.2019.101820}

\bibitem{gqm-book}
Van~Solingen, R., Basili, V., Caldiera, G., Rombach, H.D.: Goal question metric
  (gqm) approach. Encyclopedia of software engineering  (2002)

\bibitem{8120406}
{Weintrop}, D., {Shepherd}, D.C., {Francis}, P., {Franklin}, D.: Blockly goes
  to work: Block-based programming for industrial robots. In: 2017 IEEE Blocks
  and Beyond Workshop (B B). pp. 29--36 (2017).
  \doi{10.1109/BLOCKS.2017.8120406}

\bibitem{DBLP:conf/chi/WeintropASFLSF18}
Weintrop, D., Afzal, A., Salac, J., Francis, P., Li, B., Shepherd, D.C.,
  Franklin, D.: Evaluating coblox: {A} comparative study of robotics
  programming environments for adult novices. In: Mandryk, R.L., Hancock, M.,
  Perry, M., Cox, A.L. (eds.) Proceedings of the 2018 {CHI} Conference on Human
  Factors in Computing Systems, {CHI} 2018, Montreal, QC, Canada, April 21-26,
  2018. p.~366. {ACM} (2018). \doi{10.1145/3173574.3173940},
  \url{https://doi.org/10.1145/3173574.3173940}

\bibitem{DBLP:conf/mc/YigitbasHE19}
Yigitbas, E., Heind{\"{o}}rfer, J., Engels, G.: A context-aware virtual reality
  first aid training application. In: Alt, F., Bulling, A., D{\"{o}}ring, T.
  (eds.) Proc. of Mensch und Computer 2019. pp. 885--888. {GI} / {ACM} (2019)

\bibitem{DBLP:conf/interact/YigitbasJ0E19}
Yigitbas, E., Jovanovikj, I., Sauer, S., Engels, G.: On the development of
  context-aware augmented reality applications. In: Abdelnour{-}Nocera, J.L.,
  Parmaxi, A., Winckler, M., Loizides, F., Ardito, C., Bhutkar, G., Dannenmann,
  P. (eds.) Beyond Interactions - {INTERACT} 2019 {IFIP} {TC} 13 Workshops,
  Paphos, Cyprus, September 2-6, 2019, Revised Selected Papers. Lecture Notes
  in Computer Science, vol. 11930, pp. 107--120. Springer (2019).
  \doi{10.1007/978-3-030-46540-7\_11},
  \url{https://doi.org/10.1007/978-3-030-46540-7\_11}

\bibitem{DBLP:conf/vrst/YigitbasJSE20}
Yigitbas, E., Jovanovikj, I., Scholand, J., Engels, G.: {VR} training for
  warehouse management. In: Teather, R.J., Joslin, C., Stuerzlinger, W.,
  Figueroa, P., Hu, Y., Batmaz, A.U., Lee, W., Ortega, F.R. (eds.) {VRST} '20:
  26th {ACM} Symposium on Virtual Reality Software and Technology. pp.
  78:1--78:3. {ACM} (2020)

\bibitem{DBLP:journals/corr/abs-2103-10804}
Yigitbas, E., Karakaya, K., Jovanovikj, I., Engels, G.: Enhancing
  human-in-the-loop adaptive systems through digital twins and {VR} interfaces.
  CoRR  \textbf{abs/2103.10804} (2021), \url{https://arxiv.org/abs/2103.10804}

\bibitem{DBLP:conf/eics/EnesScaffolding}
Yigitbas, E., Sauer, S., Engels, G.: Using augmented reality for enhancing
  planning and measurements in the scaffolding business. In: {EICS} '21: {ACM}
  {SIGCHI} Symposium on Engineering Interactive Computing Systems, virtual,
  June 8-11, 2021. {ACM} (2021), \url{https://doi.org/10.1145/3459926.3464747}

\bibitem{DBLP:conf/mc/YigitbasTE20}
Yigitbas, E., Tejedor, C.B., Engels, G.: Experiencing and programming the
  {ENIAC} in {VR}. In: Alt, F., Schneegass, S., Hornecker, E. (eds.) Mensch und
  Computer 2020. pp. 505--506. {ACM} (2020)

\bibitem{DBLP:conf/icawst/YoshizawaW18}
Yoshizawa, Y., Watanobe, Y.: Logic error detection algorithm for novice
  programmers based on structure pattern and error degree. In: 9th
  International Conference on Awareness Science and Technology, iCAST 2018,
  Fukuoka, Japan, September 19-21, 2018. pp. 297--301. {IEEE} (2018).
  \doi{10.1109/ICAwST.2018.8517171},
  \url{https://doi.org/10.1109/ICAwST.2018.8517171}

\end{thebibliography}

\end{document}